# Language and Speech Technology for Central Kurdish Varieties

**Sina Ahmadi**[1,2], **Daban Q. Jaff**[3], **Md Mahfuz Ibn Alam**[2], **Antonios Anastasopoulos**[2,4]
[1]University of Zurich  [2]George Mason University  [3]University of Erfurt  [4]Archimedes AI Research Unit
[1]`sina.ahmadi@uzh.ch`  [2]`{malam21,antonis}@gmu.edu`  [3]`dabjaff@gmail.com`

## Abstract

Kurdish, an Indo-European language spoken by over 30 million speakers, is considered a dialect continuum and known for its diversity in language varieties. Previous studies addressing language and speech technology for Kurdish handle it in a monolithic way as a macro-language, resulting in disparities for dialects and varieties for which there are few resources and tools available. In this paper, we take a step towards developing resources for language and speech technology for varieties of Central Kurdish, creating a corpus by transcribing movies and TV series as an alternative to fieldwork. Additionally, we report the performance of machine translation, automatic speech recognition, and language identification as downstream tasks evaluated on Central Kurdish subdialects. Data and models are publicly available under an open license at https://github.com/sinaahmadi/CORDI.

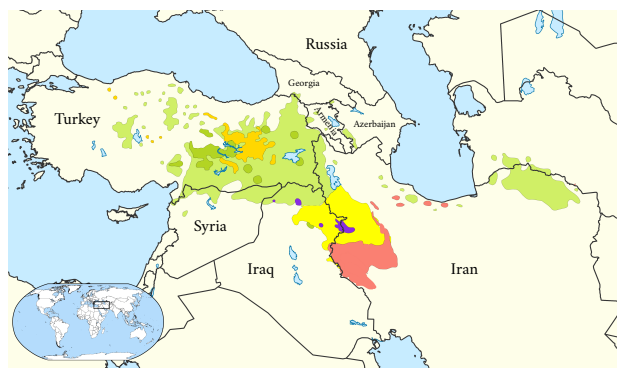

Figure 1: Geographical distribution Kurdish languages/dialects as Northern ▇, Central ▇, and Southern ▇. Zazaki ▇ and Gorani ▇ are closely related to Kurdish. ▇ refers to mixed areas. Our focus is on the dialects spoken in the yellow region. (Map created based on Ahmadi (2021)).

## 1 Introduction

In the evolution of the development of language and speech technology (LST) for a given language, varieties and dialects that are more represented in terms of data are most likely to be focused on initially. Consequently, this leads to a disparity between the speakers of various dialects of a language in using technology. For instance, despite the enormous amount of work in English, only a fraction of previous studies on dependency parsing focuses on dialects or varieties like African-American Vernacular English in comparison to Mainstream American English (Blodgett et al., 2018). Thanks to the remarkable progress in recent years in cross-lingual and low-resource natural language processing (NLP), as well as the growing accessibility of data for under-represented language varieties, many studies have gone beyond the monolithic concept of a language as in Multi-VALUE (Ziems et al., 2022). Similarly, many initiatives call for a more universal and more diverse representation of varieties and dialects in NLP (Plank, 2016; Zampieri et al., 2020).

The focus of the current paper is the Kurdish language, a less-resourced Indo-European language spoken by over 30 million speakers in Iraq, Iran, Turkey, Syria, and among the Kurdish diaspora. As a nation without a state, Kurds have been facing constant linguistic discrimination and hostile policies causing pernicious sociolinguistic effects such as lack of formal language education, engulfing Kurdish with loanwords and even linguicide (Ahmadi et al., 2023b). On the other hand, efforts have been made to standardize Kurdish throughout decades. This is particularly important for Kurdish given its extraordinary diversity in dialects and varieties, up to the point that dialectal differences can be noticed even between neighboring villages.

Corpora are essential resources for linguistic research as well as for developing language technologies. Given that the existing corpora for Kurdish rely on the journalistic register of the language, there is a dearth of resources that document vari-eties and dialects. To tackle this, we create the corpus of dialogues in Central Kurdish – CORDI. CORDI is a corpus of colloquial speech based on the dialogues in over 300 movies and series. This is the first corpus documenting colloquial speech taking a variety of dialects of Central Kurdish into account. In addition to this major contribution, we address a few downstream tasks, namely machine translation, language identification, and automatic speech recognition. We report the sub-optimal performance of state-of-the-art models on dialectal data. Although our primary focus is on dialects of Central Kurdish, our approach can be applied to other dialects and languages. In this paper, we use '(sub)dialect' and 'variety' interchangeably.

## 2 Kurdish and its Varieties

The historical heritage and strong connection to their land and culture have significantly molded the identity of the Kurdish people. Nonetheless, defining the Kurdish language and what determines its dialects has proven to be intricate and demanding, influenced by political and societal factors. In this section, we briefly lay out some of the existing classifications in Kurdish dialectology. For a more analytical description of Kurdish dialects, see Öpengin and Haig (2014); Eppler and Benedikt (2017); Belelli et al. (2019); Matras (2019).

As one of the earliest classifications of Kurdish dialects, MacKenzie (1961) proposed a general division between 'Northern' and 'Central' dialects. Since then, these along with Southern Kurdish have been widely followed in Kurdish dialectology and linguistics. Therefore, Kurdish is divided into Northern, Central, and Southern dialects, although there is some inconsistency in terminology (Edmonds, 2013; McCarus, 2013). Northern Kurdish, also known as Kurmanji, is spoken in Iran, Iraq, Turkey, and Syria, while Central Kurdish or Sorani is only spoken in Iran and Iraq, and Southern Kurdish is chiefly spoken in Kermanshah and Ilam provinces in Iran, and across the adjoining border regions of Iraq (Eberhard et al., 2022).

Although closely related languages like Zazaki and Gorani are commonly perceived as distinct from the Kurdish language, they are occasionally categorized as potential additional dialects of Kurdish as well. This perception persists despite the speakers of these languages sharing close cultural affiliations with neighboring Kurdish communities and often identifying themselves as ethnically Kurdish, as underscored by Haig and Öpengin (2014). Similarly, the classification of Laki as the southernmost variety within the Kurdish language cluster remains a subject of ongoing debate. Nevertheless, there is scholarly consensus that Lori (also spelled Luri) is unequivocally a Southwestern Iranic language, dispelling the widespread misconception that it is a variant of Kurdish, a conclusion elucidated by Anonby (2004). Nonetheless, both Lori and Laki display instances of linguistic convergence with neighboring Southern Kurdish dialects.

| Source | Dialect | Corpus |
|---|---|---|
| News | Northern | Ataman (2018) |
| | Central | Esmaili et al. (2013) |
| | Southern | Ahmadi et al. (2023b) |
| | Zazaki, Gorani | Ahmadi (2020a) |
| Fieldwork | Northern, Central | Matras (2019) |
| | Laki | Belelli (2021) |
| Textbooks | Central | Abdulrahman et al. (2019) |

Table 1: A summary of the existing corpora for Kurdish dialects. There is a lack of a more diverse dialectal material in the existing corpora.

Considering the three-branch classification, delving into the realm of Kurdish dialectology through a review of the existing literature reveals a longstanding challenge in delineating the subdialects of Kurdish and establishing clear demarcations among them. This enduring difficulty has primarily arisen from a paucity of comprehensive data. Recently, Öpengin and Haig (2014) and Matras (2019), in two different projects, respectively focus on collecting data for subdialects of Northern Kurdish (Kurmanji) and Central Kurdish. Both studies focus on features in vocabulary, phonology, and morphology and collect data through questionnaires and fieldwork. In the first study, the authors adopt a division of the Kurmanji speech zone into five regions. In the latter, the author defines layers of structural innovation zones for both Northern Kurdish and Central Kurdish dialect continuum and their geographical diffusion as follows: Kurmanji continuum with two extremities being the Southeastern Kurmanji with its epicenter in the Duhok province (Iraq) and the Western Kurmanji encompassing Muş (Turkey) and, Sorani continuum with two zones of Southern Sorani with its epicenter as Sulaymaniyah (Iraq) and Northern Sorani with its epicenter as Erbil (Iraq).

## 3 CORDI

### 3.1 Motivation

Despite the previous studies and the ongoing attempts in dialectology to create dialect maps as in Anonby et al. (2019)'s atlas of dialect in Kurdistan province of Iran and University of Kurdistan's atlas,[1] progress is hindered due to data paucity. On the other hand, the standardization process of Kurdish (Hassanpour, 1989), the lack of education of native Kurdish speakers in their mother tongue, and technological barriers, all, implicitly or explicitly, have exacerbated this paucity. According to Ahmadi et al. (2023b)'s recent survey of Kurdish corpora and language resources, summarized in Table 1, one notices that a limited number of corpora consist of colloquial or dialectal material while the majority focuses on the journalistic register of the language.

### 3.2 Data Collection

To tackle the problem of data paucity for Central Kurdish varieties, we create a text and audio corpus by transcribing movies and series. Following identifying such materials, we transcribe and synchronize the dialogues with additional speech features, such as the main characters, and the age and gender of the speakers. During the data collection phase, we noticed that the majority of audiovisual work for Central Kurdish is created by actors and actresses based in Sulaymaniyah (Iraq), Sanandaj (Iran), Erbil (Iraq), and Mahabad (Iran), and a few, from Kalar (Iraq) and Sardasht (Iran). The following local names are also generally and broadly used to refer to the dialects spoken in the regions specified in parentheses: *Babani* 'بابانی' (Sulaymaniyah, Iraq), *Ardalani* 'ئەردەڵانی' (Sanandaj, Iran), *Jafi* 'جافی' (Javanrud, Iran), *Mukriyani* or *Mukri* 'موکریانی' (Mahabad, Iran) and *Hewlêrî* 'هەولێری' (Erbil, Iraq). We opt to classify the content based on city names to avoid confusion. The cities where the films originate from are marked in Figure 2.

Considering Matras (2019)'s classification of Central Kurdish subdialects and epicenters, dialects of Sulaymaniyah, Sanandaj, and Kalar correspond to Southern Sorani and those of Erbil, Mahabad, and Sardasht correspond to Northern Sorani. As such, we believe that further comparative studies will be available thanks to our data.

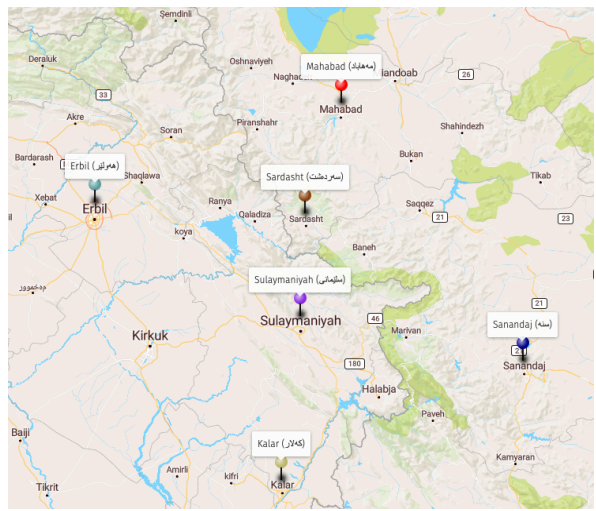

Figure 2: Cities where the selected series and films of CORDI are produced across Iran-Iraq border

### 3.3 Audio Transcription

In the context of this project, the audio transcription process was structured into the following phases:

1. **Grouping** During the initial team assembly phase, we sought to enlist the help of volunteers by putting out a call for volunteers. 55 individuals responded, primarily composed of third and fourth-year students from the Department of English Language at Faculty of Education of Koya University (Kurdistan Region). In an introductory session, volunteers were informed about the project's objectives and the selected dialects. Ultimately, 36 volunteers, comprising 26 females and 10 males, consented to participate. The entire crowd-sourcing process spanned a period of 10 months, commencing in June 2021 and concluding in April 2022.

2. **Orientation** In the orientation phase, we first focused on the technical aspects of using the Amara platform[2] for transcription. Volunteers were grouped into teams and underwent practical training sessions to familiarize themselves with the platform. They worked on short videos, lasting two to three minutes, to gain competence in using the platform. Then, the orientation was linguistically oriented, introducing volunteers to guidelines for punctuation, grammar, Kurdish orthography, corpus annotation, and identifying utterances in oral discourse. They were also instructed on how to annotate speaker age as `adult` or `child`, gender as `masculine` or `feminine`, and dialect.

---

[1] https://atlas.uok.ac.ir

[2] https://amara.org

| Variety | # Utterances | length (hours) | Ave. tokens | Ave. length (seconds) | Speaker metadata (%) |
|---|---|---|---|---|---|
| Sulaymaniyah | 115,083 | 64.44 | 9.06 | 2.39 | 78.1 |
| Sanandaj | 18,584 | 18.57 | 9.53 | 2.47 | 82.59 |
| Erbil | 39,674 | 11.2 | 7.78 | 1.68 | 89.62 |
| Mahabad | 9,410 | 4.3 | 8.45 | 2.2 | 52.28 |
| Kalar | 2,150 | 1.22 | 10.92 | 2.88 | 97.72 |
| Sardasht | 1,137 | 0.42 | 7.97 | 2.29 | 59.1 |
| Total | 186,038 | 100.15 | 8.95 | 2.32 | 76.56 |

Table 2: Basic statistics of CORDI (# refers to number).

3. **Initiation** This phase began after volunteers had been equipped with the necessary technical and linguistic knowledge. Volunteers began working on their assignments, and we monitored their progress, offering support and feedback. Social media chatting groups facilitated instant communication and guidance.

4. **Proofreading** Following the completion of tasks, the proofreading phase commenced where a group of dedicated volunteers reviewed all submissions, checking for timing, orthography, punctuation, and annotation accuracy. Once the correctness of the process was verified, volunteers were asked to upload metadata and the Amara link to a spreadsheet for further processing.

The transcription process faced several obstacles and limitations such as poor Internet connectivity, the lack of robust technological infrastructure, and the absence of financial compensation. On average, it took 40 minutes to transcribe, annotate, and synchronize 10 minutes of content, i.e. > 4000 hours overall.

### 3.4 Corpus Structure

To create the corpus, videos are downloaded from YouTube in the MP4 format using PyTube[3], then converted to audio in `.wav` format and finally, segmented to utterances according to the beginning and ending timecodes in the transcriptions (in `.srt` format) using MoviePy[4]. Each utterance is converted to `.ogg` format. The original videos are not included in the released corpus.

The corpus is structured at two levels: each individual movie or episode is segmented in a folder with a JSON file providing the text of the utterance, the audio file name, and other metadata such as the age and gender of the speaker. At a broader level, a metadata file is provided that describes each movie or episode with information such as dialect, genre, and title of the episode.

### 3.5 Corpus Statistics

CORDI contains 311 movies and episodes of the two genres of comedy and drama which are transcribed and distributed based on the dialects as follows: 222 for Sulaymaniyah (Iraq), 30 for Sanandaj (Iran), 36 for Erbil (Iraq), 18 for Mahabad (Iran), one for Sardasht (Iran) and four for Kalar (Iraq). The transcription contains 186,038 utterances among which 184,805 utterances are synchronized in text and audio. The transcriptions are cleaned and standardized based on the orthography of the language. The unsynched utterances are due to the unavailability of a video on YouTube by the time of the final preparation of the corpus. It is worth noting that the metadata regarding the speaker of each utterance was provided by the transcriber at the beginning of each utterance in parentheses. However, we found out that this was not consistent, and only a certain percentage of the utterances were properly specified. Table 2 provides this percentage along with other basic statistics of the corpus.

## 4 Comparative Dialectal Analysis

One objective of this project involves conducting comparative studies among Kurdish dialects and Central Kurdish subdialects. As part of our preliminary investigation, we carry out analyses in this section on all subdialects except for Kalar and Sardasht, as the available data for these are limited.

### 4.1 Phonetics and Phonology

A diverse set of phonemes is used in Central Kurdish subdialects. Although phonemic, Kurdish scripts do not fully support phonetic variations in writing. Notable phonetic variations are palatal af-fricates /tɕ/ and /dʑ/ in Erbil and Mahabad variants respectively instead of the post-alveolar affricates /dʒ/ and /tʃ/ in the other two subdialects. In all subdialects, velarized /sˠ/ and /zˠ/ exist as in /sˠɛwzˠ/ (سەوز, *sewz*) 'green'. Some of the notable phonological processes include the followings:

(i) reduction of /d/ to /j/ as in /ɛjɑː/ in Sulaymaniyah variant and its deletion (∅) with compensatory lengthening to /ɛ/ or /ɑ/ as in /ɛɑː/ in Sanandaj variant in comparison to '*eda*' (ئەدا, /ɛdɑː/) 'it gives' in standard Central Kurdish

(ii) palatalization of /g/ and /k/ respectively as /dʑ/ and /tʃ/ in Erbil and Mahabad variants, as in /kɛndʑeː/ vs. /kɛngeː/ (*kengê*, کەنگێ) 'when'

(iii) /b/ to /w/ alternation in Sanandaj variant as in /ɛwɑːreː/ vs. /ɛbɑːreː/ ('*ebarê*', ئەبارێ) 'it rains'

(iv) substitution in pharyngeals /ħ/ and /ʕ/ as in /ʕɛliː/ (Ali, عەلی) vs. /ħɛliː/ in Erbil and Mahabad variants

(v) substitution of liquid consonants /ɫ/ by /r/ in Erbil variant as in /dɪr/ for /dɪɫ/ ('dill', دڵ) 'heart' in Standard Central Kurdish.

---

[3] https://pytube.io
[4] https://github.com/Zulko/moviepy

Further processes are described by Asadpour (2021); Matras (2019).

| Standard | Sul. | Sanandaj | Erbil | Mahabad |
|---|---|---|---|---|
| *naw* (name) | *naw* | *naw* | *naw* | *nêw* |
| *xoş* (fine) | *xoş* | *xweş* | *xoş* | *xoş* |
| *xał* (uncle) | *xał* | *xało* | *xar* | *xał* |
| *kewtin* (fall.v) | *kewtin* | *keftin* | *kewtin* | *kewtin* |
| *xwên* (blood) | *xwên* | *xwên* | *xîn* | *xên* |
| *ziman* (language) | *ziman* | *ziwan* | *ziman* | *ziman* |
| *deçim* (I go) | *eçim* | *eçim* | *deçim* | *deçim* |
| *jinan* (women) | *jinan* | *jingel* | *jinan* | *jinan* |

Table 3: Phonological and morphological differences across words in Central Kurdish varieties.

### 4.2 Morphology and Syntax

Despite the similarity of Central Kurdish subdialects in morphosyntax, there are a few remarkable variations across the selected subdialects. Being a split-ergative language, all the subdialects exhibit a difference in marking transitive verbs in the past tenses. The Sanandaj subdialect, in comparison to the other subdialects, follows a relatively different pattern in the alignment of subject-object, both at the word level and the sentence level as in بردمیان (*birdimyan*) 'they brought me' in Sanandaj subdialect versus بردیانم (*birdyanim*) where the patient marker -*im* appears in a different order. Another notable variation is due to marking oblique cases in the Erbil and Mahabad subdialects unlike the other two. There are a few syntactic variations such as the tendency of adpositions in verbal compounds in the Sanandaj subdialect to appear after the verb as in بۆ (*bo*) 'for' in هاوردم بۆت (*hawirdim bot*) 'I brought it to you.2sɢ', unlike the other subdialects where بۆم هێنای (*bom hênay*) is used.

Table 4 provides a comparison of the most frequent bound morphemes in the selected subdialects with a few examples in Table 3. Some of the morphemes are used across all the subdialects with varying frequency. A more comprehensive comparison is provided in (Ahmadi, 2021).

### 4.3 Lexicon

Lexical variations across Kurdish varieties are remarkable. For instance, جووشک (*cûşik*) 'hedgehog' in Mahabad variant versus ژوژوو (*jûjû*), ژیشک (*jîşik*) and ژووژک (*jûjik*) respectively in Sanandaj, Erbil and Sulaymaniyah variants. Some of such variations are due to phonological alternations of words of identical cognates. On the other hand, a considerable number of words have different cognates as in وەوی (*wewî*) 'bride' in Sanandaj (probably historically borrowed in contact with Gorani) in comparison to بووک (*bûk*) in the other subdialects. Nevertheless, a considerable factor in lexical variation is due to loanwords and terminologies that are influenced by the dominant language in the administration of each region. As such, loanwords in Sanandaj and Mahabad variants such as ئاسانسۆر (*asansor*, 'elevator') from Persian←French '*ascenseur*' are more likely to look alike under the influence of Persian, analogous to the other two subdialects under the influence of Arabic, e.g. مەسعەد (*mes'ed*) from Arabic مصعد.

### 4.4 Standard Central Kurdish

Although still vague, Standard Central Kurdish broadly refers to the language used in the press and the media. It is believed that among the subdialects, that of Sulaymaniyah is widely adopted as a standard variant of Central Kurdish (Thackston, 2006). That said, Standard Central Kurdish does not completely conform with a specific subdialect. For instance, the suffix وانێ (-*ewanê*) in هاتمەوانێ (*hatimewanê*) 'I came back' used in Sulaymaniyah variant is not used in the standard variant.

Considering morphosyntactic and lexical variations, we estimate the similarity of Standard Cen-tral Kurdish and that of the subdialects by creating a list of 60 words among the most frequent morphemes and a list of 145 words in the general vocabulary. Comparing the overlap of the variants in this list indicates that Central Kurdish has indeed a standardized variant which is not identical to any of the variants. As illustrated in Figure 3, the Sulaymaniyah variant is the closest one to Standard Central Kurdish while the Sanandaj variant is the least similar one. Similarly, the Mahabad variant shows more similarity with the Erbil variant while Sulaymaniyah and Sanandaj variants are closer. Interestingly, this is in line with Matras (2019)'s epicenters of Northern and Southern Central Kurdish.

| Part-of-speech | | | Southern Central Kurdish | | Northern Central Kurdish | |
|---|---|---|---|---|---|---|
| | | | Sulaymaniyah | Sanandaj | Erbil | Mahabad |
| Noun | INDF | SG | -êk ( ێک ) | -êk | -ek ( ەک ) | -êk |
| | | PL | -an ( ان ) | -gel ( گەل ) | -an | -an |
| | DEF | SG | -eke ( ەکە ), -e ( ە ) | -eke, -e | -eke, -e | -eke, -e |
| | | PL | -ekan ( ەکان ) | -ekan | -ekan | -ekan |
| | DEM | SG/PL | -e ( ە ) | -e | -e | -e |
| | OBL | | - | - | -î ( ی ), -ê ( ێ ) | -î, -ê |
| | IZAFE | | -î ( ی ), -e ( ە ) | ∅, -e | -î, -e | -î, -e |
| Verb | INF | | -in ( ن ) | -in | -in | -in |
| | PROG | | e- ( ئە ) | e- | de- ( دە ) | de- |
| | SBJV | | bi- ( بـ ) | bi- | bi- | bi- |
| | NEG | | ne- ( نە ), na- ( نا ) | ne-, na- | ne-, na- | ne-, na- |
| | suffix -ewe ( ەوە ) | | -ewe, -ewanê ( ەوانێ ) | -ew ( ەو ) | -ewe | -ewe |
| | Adverbial -ê ( ێ ) | | ✓ | ✗ | ✓ | ✓ |
| | Clitic =îş ( یش ) | | ✓ | ✓ | ✓ | ✓ |
| Adjective | COMP | | -tir ( تر ) | -tir | -tir | -tir |
| | SUP | | -tirîn ( ترین ) | -tirîn | -tirîn | -tirîn |

Table 4: A comparison of the most frequent bound morphemes in Central Kurdish varieties.

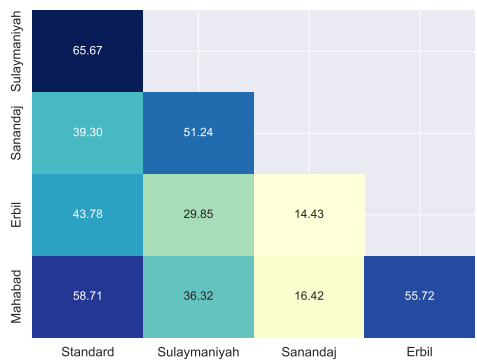

Figure 3: The similarity (percentage) of the selected subdialects in CORDI using a set of morphosyntactic and lexical variations. Standard Central Kurdish is indeed a variant on its own and shares the most similarity with the Sulaymaniyah variant and the least, with the Sanandaj variant.

## 5 Experiments

In addition to the dialectal linguistic analysis, we use our corpus to build and evaluate models for downstream applications, namely automatic speech recognition, machine translation, and language identification. We expect that pre-trained models perform poorly when evaluated on dialect data in comparison to Standard Central Kurdish.

### 5.1 Machine Translation

Machine translation (MT) is one of the most important NLP applications and has been previously addressed for Kurdish, mainly for Northern and Central Kurdish (Ahmadi et al., 2022; Ahmadi and Masoud, 2020; Amini et al., 2021). Similarly, Northern and Central Kurdish are currently supported by Google Translate[5] and Bing Microsoft Translator.[6]

Using our corpus, we aim to demonstrate how the existing models perform on Central Kurdish subdialects, as the robustness of neural MT (NMT) systems is known to be affected by dialectal variations (Vaibhav et al., 2019).

---
[5] https://translate.google.com
[6] https://www.bing.com/translator

|  | English → Central Kurdish Variety | | | | Central Kurdish Variety → English | | | |
|---|---|---|---|---|---|---|---|---|
|  | NLLB | | Google | | NLLB | | Google | |
|  | Baseline | postprocess | Baseline | postprocess | Baseline | preprocess | Baseline | preprocess |
| Standard | 1.5 (25.5) |  | 3.5 (33.6) |  | 11.4 (28.6) |  | 22.4 (42.9) |  |
| Sulaymaniyah | 1.5 (25.3) | 1.6 (26) | 3.3 (33.2) | 3.5 (33.9) | 10.6 (27.9) | 11.2 (28.5) | 22.7 (43.3) | 25 (43.2) |
| Sanandaj | 0.6 (19.9) | 0.7 (22) | 1.1 (24.7) | 2.1 (27.3) | 6.5 (21.6) | 7.4 (22.8) | 15.6 (35.9) | 17.4 (35.9) |
| Erbil | 1.2 (24.5) | 1.3 (25.3) | 3.1 (31.2) | 3.4 (31.7) | 10.4 (27.6) | 11.4 (28.5) | 21.2 (41.9) | 23.3 (42.1) |
| Mahabad | 0.6 (22.5) | 0.8 (23.9) | 2.5 (29.3) | 3.5 (30.8) | 7.1 (24) | 8.6 (25.2) | 18.6 (39) | 19.8 (38.8) |

Table 5: Evaluation of pretrained NMT models for translation of English into Central Kurdish (EN→CKB) and vice versa (CKB→EN) using BLEU scores with chrF2 in parentheses (higher is better). Even though the models struggle to translate dialects effectively with Google Translate outperforming NLLB, the postprocess and preprocess procedures ameliorate the translation quality.

**Parallel Corpus** For the evaluation of MT systems, we first create a gold-standard parallel corpus based on CORDI. To do so, we randomly selected sentences from the utterances of four subdialects of Sulaymaniyah, Erbil, Sanandaj, and Mahabad in our corpus. These 300 sentences are then translated by native speakers into other dialects, Standard Central Kurdish as well as into English. To reduce the influence of writing differences, we normalize the script of the sentences based on Kurdish orthography using KLPT (Ahmadi, 2020b). This corpus is included in CODET (Alam et al., 2023).

**Setup** We use Meta AI's No Language Left Behind (NLLB) (Costa-jussà et al., 2022) model on HuggingFace[7] which is trained on parallel multilingual data from a variety of sources to translate into English and evaluate using BLEU score and chrF2 in SacreBLEU (Post, 2018). The chrF2 score complements BLEU by providing a more nuanced evaluation that takes into account longer character $n$-grams, thereby offering a more robust assessment of translation quality. In addition to NLLB, we evaluate the translations of Google Translate through their web interfaces.

**Dialectalization and Standardization** As an additional step, we add a set of rules to convert a text from Standard Central Kurdish into one of the selected dialects, i.e. dialectalization. We carry out the reverse task of standardization, i.e. conversion from a dialect to the standard, as well. Our objective is to assess the effect of such rules in improving the performance of NMT systems and consequently, the difficulty of dialect translation given the pretrained models.

Given a text in Standard Central Kurdish, the standardization process consists of the following steps:

1. **Apply morphosyntactic rules** Using KLPT's morphological analyzer, analyze a given word, segment it into composing morphemes, and finally, replace those defined in the rules with their dialectal equivalents. Our rules include changing article marking suffixes as in -êk→-ek for Erbil variant, verbal progressive prefixes as in e-→de- in Sanandaj and Sulaymaniyah variants and discontinuous circumpositions. Our rules include the variations in Table 4 and §4.2.

2. **Map Vocabulary** Map the frequent free morphemes such as adpositions and numerals and also, general vocabulary defined for the dialects. Our mapping contains 145 general vocabulary words and 60 free morphemes (including inflected ones) also described in §4.4.

3. **Replace Terminology** As the terminology varies across the border (Iran-Iraq), terms and loanwords employed in dialects spoken under the two administrations are not identical. As the final step, we replace such words. Our mapping contains 58 such terms and loanwords.

The standardization process follows the same procedure reversely to a dialectal sentence to look like a standard one. We evaluate the impact of these processes through dialectalization of translation hypothesis (postprocess) and standardization of dialectal input (preprocess). As such, we define baseline setup where these additional procedures are not taken into account.

**Results** Our experimental results are provided in Table 5 using BLEU scores along with chrF2 in parentheses. While the baseline models perform poorly in EN→CKB using both NLLB and Google, it has higher scores in CKB→EN. Moreover, Google Translate demonstrates increased resilience to dialectal variations, surpassing the established baseline. Interestingly, our `postprocess` and `preprocess` approaches yield modest quality improvements, albeit with a discernible constraint, underscoring the intricacies inherent in the deployment of an all-encompassing rule-based strategy for addressing nonstandard textual content. We believe that the use of colloquial language in both source and target languages exacerbates the translation challenge.

---

[7]The `nllb-200-distilled-600M` variant.

| Data | CV-Scratch | CV-PT-en | CORDI-Scratch | CORDI-PT-en | CORDI-PT-CV |
| --- | --- | --- | --- | --- | --- |
| Sulaymaniyah | 125.42 | 112.11 | **58.56** | 62.9 | 60.97 |
| Sanandaj | 131.7 | 111.68 | **58.98** | 67.84 | 60.84 |
| Erbil | 117.67 | 114.4 | 72 | 74.96 | **71.7** |
| Mahabad | 128.43 | 116.03 | **66.88** | 68.32 | 69.69 |
| CommonVoice | 148.99 | **85.6** | 170.14 | 171.86 | 236.51 |

Table 6: Experimental results of ASR using different models and test sets based on WER (lower is better) show that the task is challenging for Central Kurdish subdialects. The lowest WER scores are in boldface.

## 5.2 Automatic Speech Recognition

Automatic speech recognition (ASR) is a technology that transforms spoken language into written text. Although the previous studies address ASR for Kurdish, as in building a speech corpus and pronunciation lexicon (Veisi et al., 2022) or dialect recognition using speaker embeddings (Amani et al., 2021), creating technologies for sub-dialects of Central Kurdish has not been discussed previously. Similarly, the CommonVoice corpus (Ardila et al., 2020b) only focuses on Standard Central Kurdish, even though the pronunciation of the recorders may vary based on the recorders.

To evaluate the performance of the existing dataset, we run experiments targeting specific sub-dialects in our corpus. Additionally, we train a model based on our dataset and evaluate it as well. The evaluation is conducted using the Word Error Rate (WER) metric, which calculates the proportion of errors in the ASR output relative to the total number of words in the reference transcript.

**Dataset** To create a dataset, we first filter out any utterance with a length of less than 0.2 seconds. We then split our dataset into train, test, and validation sets. Our dataset consists of four dialects of Erbil, Mahabad, Sulaymaniyah, and Sanandaj. Due to a limited number of instances, we randomly chose 500 utterances as the test set for Mehabad and Sanandaj dialects, and for Erbil and Sulaymaniyah, we randomly chose 2000 utterances as the test set. The remaining utterances were used as train and validation. 10% of the utterances were randomly chosen as the validation set.

**Setup** We use Fairseq (Ott et al., 2019) to train our ASR system. The model architecture is *s2t_transformer_s*, which is trained for 100,000 updates using Adam optimizer with a learning rate of 2e-3 and dropout of 0.3. The target side vocabulary was trained using SentencePiece tokenization (Kudo, 2018). We have kept the vocab size fixed at 4,000. The encoder weights are loaded for some models with a previously trained ASR encoder.

**Results** Table 6 shows the result of our ASR experiments. We create five models and evaluate these models with five test sets, i.e. the selected four dialects and that of CommonVoice. We create two models using just the CommonVoice train set as a baseline. One model was trained from scratch and another was trained by loading a pre-trained ASR model's encoder. We refer to these two models as `CV-Scratch` and `CV-PT-en`, respectively. The pre-trained model was trained using the same setup as above, but the training dataset for this model was the English ASR data of the `en-de` language pair of MustC dataset (Di Gangi et al., 2019). Our results indicate that the pre-trained encoder, even if from different languages, yields a boost in WER when tested on the CV dataset. WER is greater than 100 in both baseline models.

Following this, we create three models using the train set of CORDI. The first model is trained from scratch without any pre-trained model. We refer to this as `CORDI-Scratch`. Among the four dialects, this model yields the best score in three dialects: Mahabad, Sulaymaniyah, and Sanandaj. For the other two, we use the weight of two different pre-trained models' encoders. We load the same pre-trained English ASR model as in the first model. For the second model, we load the best baseline model's encoder, `CV-PT-en`, based on the result of the CommonVoice test set. The results of these two models show that if the pre-trained models are available, using the pre-trained model trained in the same language gives a higher boost even if the pre-trained model was not good. We get the best results for the remaining dialect of Erbil from the model `CORDI-PT-CV`.

One interesting finding is that models trained on our data perform worse when tested on the CommonVoice test set, giving us the impression that the CommonVoice dataset may be substantially different from our selected dialects. Another interesting finding is that if a dialect is challenging, the pre-trained model helps more than training from scratch. According to the WER scores where `CORDI-PT-CV` is the best model, the scores are more than 70 WER but where CORDI-Scratch is the best model, WER is less than 67.

### 5.3 Language Identification

Language identification (LID) involves the process of determining the language in which a given sentence is written. This task holds significance in various NLP applications. While previous work has tackled language identification for specific scripts or for Kurdish dialects as in (Ahmadi et al., 2023b), there remains a need to encompass subdialects.

**Dataset** We create a gold-standard dataset for evaluating the identification of Central Kurdish subdialects. To do so, we rely on the transcriptions of the utterances in CORDI of the following dialects along with their corresponding labels: Sulaymaniyah (`ckb-slm`), Sanandaj (`ckb-snn`), Mahabad (`ckb-mhb`), Erbil (`ckb-hwl`) and Kalar (`ckb-klr`). We rely on 20% of the transcripts to create a test set and 80% for the train set. To tackle class imbalance in the case of Kalar and Mahabd dialects, we upsample the utterances in the train set. Overall, the train and test sets respectively contain 147,597 and 26,500 instances.

**Setup** As the baseline system, we evaluate the pre-trained language identification model of fast-Text (Bojanowski et al., 2017) which can recognize 176 languages including Northern Kurdish, Central Kurdish, and Zazaki, respectively with `kmr`, `ckb` and `diq` class labels. We train our model using fastText with the following hyper-parameters: 25 epochs, word vectors of size 64, a minimum and maximum length of char $n$-gram of 2 to 6, a learning rate of 1.0, and hierarchical softmax as the loss function. We create the classifiers at three levels:

(i) **Language** where the labels of all sentences in different varieties (Kurmanji, Zazaki, Gorani and Southern Kurdish) are unified based on the language code only. This helps to compare the performance of some of the existing identifiers, namely fastText, PALI (Ahmadi et al., 2023a) and KurdishLID (Ahmadi et al., 2023b). Both PALI and KurdishLID detect Northern, Central and Southern Kurdish along with Gorani and Zazaki among other languages. Our model is not trained at this level.

(ii) **Dialect** where sentences are classified based on the dialects

(iii) **Subdialect** where sentences are classified based on the subdialects of Central Kurdish, e.g. `ckb-mhb` and `ckb-snn` for Mahabad and Sanandaj subdialects, respectively. Expectedly, other models cannot identify these as they are not trained on.

| Model / Level | Language | Dialect | Subdialect |
|---|---|---|---|
| fastText | 0.94 | 0.94 | 0 |
| PALI | 0.958 | 0.07 | 0 |
| KurdishLID | 0.986 | 0.19 | 0 |
| Ours | 0 | 1 | 0.76 |

Table 7: Experimental results for language identification at the language, dialect and subdialect levels using different models based on F1 score of the first prediction (higher is better).

**Results** Table 7 presents the LID results with a heatmap in Figure 4. LID at the language level reveals to be a relatively easy task for all the models, including fastText. This is expected given that our test set contains Kurdish sentences only and all the models are trained on Kurdish data with Central Kurdish written in the Perso-Arabic script. On the other hand, models perform less efficiently in detecting dialects indicating that the models confuse sentences in subdialects with other varieties, notably Southern Kurdish and Gorani. Even though PALI and KurdishLID are trained at a more fine-grained level in comparison to fastText, the latter performs better. Regarding subdialects, none of the models except ours are trained for subdialects, as such, they cannot differentiate subdialects. The F1 score of 0.76 of our model indicates the difficulty of detection of the selected varieties.

|  | Sanandaj | Sulaymaniyah | Mahabad | Erbil | Kalar |
|---|---|---|---|---|---|
| Sanandaj | 4331 | 85 | 120 | 39 | 80 |
| Sulaymaniyah | 438 | 9245 | 1378 | 778 | 878 |
| Mahabad | 63 | 141 | 2371 | 135 | 42 |
| Erbil | 146 | 461 | 1082 | 4021 | 193 |
| Kalar | 22 | 68 | 49 | 27 | 307 |

Figure 4: Classification of Central Kurdish subdialects. Our model mismatches the classification of variants of Erbil and Mahbad frequently and that of Sulaymaniyah across all the variants.

## 6 Conclusion

This paper presents a novel approach for creating an audio and text corpus for Central Kurdish subdialects called CORDI. Relying on the utterances in movies and series, our approach circumvents problems of lack of financial support and expertise to create a corpus in a more conventional way, say fieldwork. CORDI contains 186,038 utterances in six subdialects of major Kurdish cities, namely Sulaymaniyah, Sanandaj, Kalar, Mahabad, Erbil, and Sardasht. Our comprehensive experiments on downstream tasks reveal that the existing models exhibit suboptimal performance when subjected to evaluation on subdialects. Consequently, it becomes evident that additional advancements are imperative to address nonstandard NLP effectively. We believe that our resources along with the other existing ones like KASET (Delgado et al., 2024) pave the way for further advances in Kurdish LST.

**Limitations** The phonetic diversity of the selected subdialects transpired to be one of the main challenges in transcribing the utterances. This could be handled by adding phonetic transcriptions to the corpus. Furthermore, the speakers in the targeted movies or series, may not be native in the dialect of the particular city. As such, it is possible to have a mix of dialects in the corpus. Given the source of the data, it is evident that the audio quality of the utterances in CORDI is not comparable with material recorded in a studio. Another limitation of our work is a small proportion of the utterance without metadata for the speaker. This excludes a portion of the corpus for linguistic studies based on gender and age.

In the future, it is advisable to extend the ongoing development of LST tailored for Kurdish subdialects, while concurrently harnessing existing models, resources, and tools for this purpose. Although comparative studies on Kurdish dialects to create dialectal dictionaries have been carried out since more than a century ago (see Morgan (1894)), there is currently a lack of a comprehensive electronic dictionary of subdialects that should be tackled in the future. Finally, creating benchmarks is needed for cross-lingual and cross-dialectal evaluation purposes.

**Ethics Statement** The material utilized in constructing our corpus is publicly accessible, ensuring compliance with data privacy regulations. To address concerns regarding copyright issues, we note that the selected movies and series were broadcast in local public channels in Iraqi Kurdistan with no restrictive license, hence freely available. We only decided to download them on YouTube as it was easier from a technical point of view. Although we made every effort to eliminate any personally identifiable information and safeguard the confidentiality and anonymity of individuals in the transcriptions, it is conceivable that certain selected utterances may contain sensitive or offensive content. The challenge of filtering such material is exacerbated by the absence of effective NLP tools. The transcription, annotation, and translation of the data, as detailed in sections §3.3 and §5.1, was carried out by volunteers who provided their consent. By adhering to these ethical principles, our aim was to conduct the study in a responsible and considerate manner.

## Acknowledgements

We express our heartfelt gratitude to the volunteers at Koya University who actively participated in the transcription and annotation tasks spanning a period of ten months from June 2021 to April 2022. Many low-resourced languages face financial constraints and Kurdish is regrettably no exception. Nevertheless, the 36 individuals mentioned below (in alphabetical order) showcase how passion for a language can overcome obstacles *despite not being remunerated monetarily*:

**Niyan** Abdulla Omer, **Sharmin** Ahmadi, **Shnya** Aram Ahmad, **Muhammad** Aram Jalal, **Zaytwn** Awny Sabir, **Lavin** Azwar Omar, **Shnyar** Bakhtyar Karim, **Sazan** Barzani Ali, **Rayan** Bestun Abdulla, **Roshna** Bestun Abdulla, **Halala** Edres Omer, **Elaf** Farhad Muhammad, **Sima** Farhad Qadr, **Ibrahem** Ismail Nadr, **Chnar** Kamal Sleman, **Muhamad** Kamaran Ahmad, **Raman** Kazm Hamad, **Burhan** Luqman Khursheed, **Shaima** Mikaeel Esmaeel, **Lavan** Muhammad Smail, **Dween** Muhammed Jamal, **Hana** Muhammed Rashid, **Amen** Muhseen Nasr, **Bryar** Murshid Mustafa, **Rayan** Mzafar Tofiq, **Taban** Omar Mohamad, **Nian** Qasim Jaff, **Dilan** Raza Nadr, **Razaw** S. Bor, **Soma** Salam Arif, **Zulaykha** Samad Abdulla, **Awdang** Saman Abdullqahar, **Eman** Sardar Hamed, **Sakar** Star Omar, **Nawa** Taha Yasin, **Triska** Zrar Mawlood. **Muhammad** Aziz, **Harman** Hameed and **Xaliss** Jamal kindly contributed to the translation of sentences of the Erbil variant in the parallel corpus as well.

Sina Ahmadi acknowledges support of the Swiss National Science Foundation (MUTAMUR; no. 213976). Daban Q. Jaff wishes to thank the *Deutscher Akademischer Austauschdienst* for his doctoral research grant (ref. no. 91826573). Antonios Anastasopoulos and Md Mahfuz Ibn Alam were supported by NSF awards BCS-2109578 and IIS-2125466 and, a Sponsored Research Award from Meta. The authors are also grateful to the anonymous reviewers for their feedback, as well as the Department of English Language at Faculty of Education of Koya University and the Office of Research Computing at George Mason University, where all computational experiments were conducted.